\documentclass[conference]{IEEEtran}
\IEEEoverridecommandlockouts

\usepackage{amsmath,amssymb,amsfonts}
\usepackage{graphicx}
\usepackage{booktabs}
\usepackage{array}
\usepackage{makecell}
\usepackage{url}
\usepackage[hidelinks]{hyperref}

\begin{document}

\title{Learn the Interface, Not the Implementation:\\
Output Representation Dominates Training Regime\\
in LLM-Based RTL Generation
\thanks{$^{\dagger}$Corresponding author. This work has been submitted to the IEEE for possible publication. Copyright may be transferred without notice, after which this version may no longer be accessible.}}

\author{
\IEEEauthorblockN{Siddharth Patel\textsuperscript{\dag}}
\IEEEauthorblockA{\textit{Dept. of Electronics Engineering}\\
\textit{Shiv Nadar Institution of Eminence}\\
Delhi-NCR, India\\
sp839@snu.edu.in}
\and
\IEEEauthorblockN{Vikram Singh} \IEEEauthorblockA{ \textit{STMicroelectronics}\\ Greater Noida, India\\ vikram.singh@st.com }
\and
\IEEEauthorblockN{Siddharth Savyasachi Malu}
\IEEEauthorblockA{
\textit{AI Center of Excellence}\\
\textit{Shiv Nadar Institution of Eminence}\\
Delhi-NCR, India\\
siddharth.malu@snu.edu.in
}
\and
\IEEEauthorblockN{Rohit Singh}
\IEEEauthorblockA{\textit{Dept. of Electronics Engineering}\\
\textit{Shiv Nadar Institution of Eminence}\\
Delhi-NCR, India\\
rohit.singh@snu.edu.in}
}

\maketitle

\begin{abstract}
Large-language-model register-transfer-level (RTL) generation has concentrated on
\emph{how to train} the model--fine-tuning, continued pre-training, reinforcement
learning from a verifier. We argue this is the second-order question. On a bounded,
formally-specifiable task--generating single-error-correcting, double-error-detecting
(SECDED) ECC memory wrappers--we hold the task, data, and correctness oracle fixed
and vary two axes independently: the \emph{training regime} (seven settings) and the
\emph{output representation} (four levels, from fully-expanded flat Verilog to an
eight-field configuration expanded by a deterministic generator). Because a reference
generator exists, correctness is \emph{decided} by formal equivalence over a
$10^{7}$-point design space, not sampled from a hand-written testbench. Two results
follow. \textbf{(1)~Representation dominates:} every representation near raw RTL
collapses to a $\le$0.17 pass rate, while the configuration reaches 0.83 with no
training at all--and once the model emits the configuration, \emph{every} training
regime lands in a 0.89--1.00 band. \textbf{(2)~Raw RTL hits a hard output-length
wall:} a 512-bit design is 44{,}237 tokens of flat Verilog against 69 tokens of
configuration, so raw generation becomes \emph{infeasible} above 32-bit data words
rather than merely inaccurate. A sub-50M-parameter from-scratch model plus the
verified generator reaches 0.89 formal-equivalence (against a 32B fine-tune's 1.00)
at $\sim$1600$\times$ lower inference energy per \emph{correct} design, and inference
energy dominates amortized training. The lesson is to learn the interface, not the
implementation.
\end{abstract}

\begin{IEEEkeywords}
LLM for RTL, hardware generation, error-correcting codes, energy-efficient ML,
program synthesis, neurosymbolic EDA, machine learning for design automation.
\end{IEEEkeywords}

\section{Introduction}
\IEEEPARstart{T}{o} emit the Verilog for a 512-bit ECC-protected memory, a language
model must produce roughly 44{,}237 tokens of flat parity logic. To specify the
\emph{same} hardware as an eight-field configuration---which a deterministic generator
expands into that exact Verilog in milliseconds---takes 69. The two descriptions are
equivalent; the gap between them is 640$\times$, and this paper is about that gap. The
prevailing question in LLM-based hardware generation is \emph{how to train} the
model---fine-tune it, pre-train it, reward it from a verifier. We show that a prior and
larger question---\emph{what should the model output?}---dominates it, deciding whether
generation is even feasible, how often it is correct, and what it costs in energy. The
incumbent a learned generator must beat is not another model but a deterministic tool
that already emits every design correctly in six seconds; the only way a model earns its
place is by writing less.

We make this precise. Fix a task with three properties that make it an unusually
honest benchmark: it is \emph{bounded} (a finite, enumerable parameter space),
\emph{parameterised} (structure repeats, so it is learnable), and \emph{formally
checkable} (a reference exists, so correctness is decidable rather than estimated).
SECDED ECC memory-wrapper generation has all three. Every point in the space has a
deterministic reference output, so we can ask of any model-generated design a question
most RTL-generation work cannot answer cleanly: \emph{is it formally equivalent to the
reference?}

\begin{figure*}[t]\centering
\includegraphics[width=0.94\textwidth]{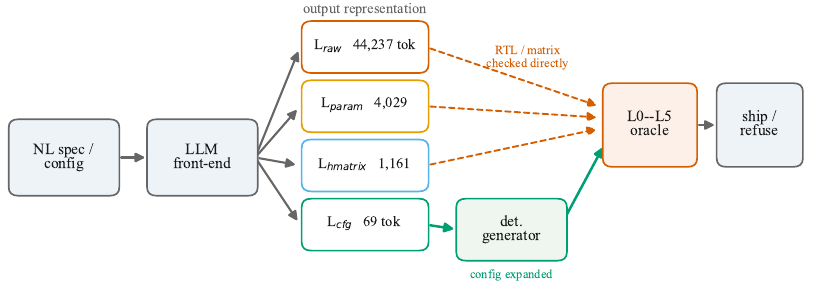}
\caption{The pipeline and the representation ladder. A natural-language spec or
configuration enters the LLM front-end, which emits one of four output
representations---shown with their 512-bit token counts. Only the configuration
(L\textsubscript{cfg}) is \emph{expanded} by a deterministic generator; the raw, parameterised,
and parity-matrix forms are emitted as implementation and checked directly against the reference by
a five-level oracle (L0--L5), then shipped or refused. The two longest representations exceed a
practical single-response generation budget---the output-length wall.}
\label{fig:overview}
\end{figure*}

On this substrate we run a two-axis study. The first axis is the \textbf{training
regime}: zero-shot, few-shot in-context learning, retrieval-augmented generation, LoRA
and full supervised fine-tuning, continued pre-training then fine-tuning, and
reinforcement learning from the formal oracle. This axis is where the literature lives.
The second axis, which we argue matters more, is the \textbf{output representation}: the
same design can be emitted as (L\textsubscript{raw}) flat Verilog, (L\textsubscript{param})
parameterised Verilog with generate loops, (L\textsubscript{hmatrix}) a parity-check
matrix plus wrapper options, or (L\textsubscript{cfg}) an eight-field configuration that a
deterministic back-end expands into RTL. These four describe \emph{the same hardware} and
span three orders of magnitude in generation length. Holding the model and recipe constant
and varying only the representation isolates its causal effect (Fig.~\ref{fig:overview}).

\noindent\textbf{Contributions.}
\begin{itemize}
\item \textbf{An oracle-complete benchmark.} Because a deterministic reference generator
exists, every point in a $10^{7}$-reachable design space has a machine-verified reference
and formal equivalence is \emph{decidable}. To our knowledge this is the first
RTL-generation benchmark where correctness is decided rather than sampled from a small
hand-written test set.
\item \textbf{A controlled two-axis study}---seven training regimes $\times$ four output
representations, one task, one dataset, one oracle---that isolates representation from
training regime and shows the former dominates.
\item \textbf{The first energy-accounted RTL-LLM study}, reported in the unit a deployment
cares about (joules per \emph{correct} design), together with a data-leakage result and a
width-disjoint split protocol that removes it.
\item \textbf{A specification-compression result}: a sub-50M-parameter model, backed by the
verified generator, reaches 0.89 formal-equivalence (against a 32B fine-tune's 1.00) at
$\sim$1600$\times$ lower inference energy per correct design---evidence that for bounded,
formally-specifiable tasks one should learn the \emph{interface}, not the
\emph{implementation}.
\end{itemize}

\section{Related Work}\label{sec:related}
\textbf{LLMs for RTL.} A line of work benchmarks and improves language-model generation of
Verilog: evaluation suites~\cite{verilogeval,rtllm}, data- and fine-tuning-centric
methods~\cite{verigen,rtlcoder,chipnemo}, correctness- and repair-oriented
approaches~\cite{craftrtl,origen}, and preference- or RL-based tuning~\cite{betterv,rl4rtl}.
These works benchmark \emph{models}: they hold the output (Verilog) fixed and vary the model
or its training. Our study is orthogonal---we hold the model family fixed and vary the
\emph{output representation}, and show that axis dominates.
\textbf{Program synthesis.} That generating into a constrained intermediate representation
and expanding deterministically beats generating the final artifact directly is old in
program synthesis and grammar-constrained decoding~\cite{synth1,grammar1}; we give a
quantitative, energy-accounted instance for hardware and identify the mechanism---an
output-length wall---that makes it not merely better but necessary at scale.
\textbf{Energy accounting.} Methodology for measuring ML training and inference
energy~\cite{strubell,patterson,mlperfpower,zeus} motivates our instrumentation; to our
knowledge no prior RTL-generation study reports energy, and we report it per \emph{correct}
design.

\section{Task and Benchmark}\label{sec:benchmark}
\textbf{SECDED wrappers.} A SECDED code augments a $k$-bit data word with $r$ check bits so
that any single-bit error is corrected and any double-bit error detected; we use the Hsiao
odd-weight-column construction, which minimizes encoder/decoder XOR depth. The wrapper composes
this datapath with a memory macro and options for port configuration, diagnostic readout, debug
exposure, ECC-bypass, and byte-write. Byte writes require a read-modify-write, which is only
meaningful on a single-port memory, a design constraint we adopt and justify, not an arbitrary
restriction.

\textbf{Design space and the leakage trap.} The deployment space is the enumerable product of
depth, width, port type, and four options subject to the byte-write rule: 3{,}584 valid
configurations. A structural fact drives everything: the encoder and decoder depend on the data
width \emph{alone}, so across all 3{,}584 configurations there are only eight distinct encoders
and eight distinct decoders -- all combinatorial variety lives in the wrapper. Hence a config that
shares its width with any training config shares its encoder \emph{file byte-for-byte}, and a
random config-level split leaks labels. We measure this: under config-disjoint splits the
train--test near-duplicate rate reaches 0.65--1.00 depending on width overlap, and it rises
monotonically with corpus size (the lowest split grows from 0.53 at 2{,}000 samples to 0.65 at
6{,}000), so these are a conservative lower bound. We therefore adopt \emph{width-disjoint} splits
for every generalization claim: test widths never appear in training in any representation, so a
held-out design's encoder was never seen in any form. Because the reference generator accepts
arbitrary widths and depths, it is a free, perfectly-labelled data source over $\sim$$10^{7}$
verifiable configurations -- which is what lets a from-scratch model be trained at all---while the
deployment grid serves as a held-out, in-distribution test set.

\textbf{The oracle harness.} Correctness is decided by a five-level harness, each a strict gate.
L0~lints; L1~elaborates; L2~runs a self-checking testbench with exhaustive single-bit fault
injection and random double-bit injection; L3~formally checks equivalence against the reference;
L4~synthesizes and reports a PPA delta; L5~checks that illegal configurations are refused. Each
level is validated on the reference itself before it scores models, so a passing score is
trustworthy; L3/L4 use an open equivalence/synthesis flow (Yosys~\cite{synth1}).

\textbf{The representation ladder.} Table~\ref{tab:ladder} defines the four output languages and
their generation lengths for the widest design: the same task, spanning three orders of magnitude
in length, with a full deterministic expander only at the shortest rung.

\begin{table}[t]
\caption{The four output representations (512-bit design).}
\label{tab:ladder}
\centering
\begin{tabular}{llr l}
\toprule
Level & Output & Tokens & Back-end \\
\midrule
L\textsubscript{raw}     & flat expanded Verilog         & $\approx$44{,}237 & none \\
L\textsubscript{param}   & parameterised, generate loops & $\approx$4{,}029  & none \\
L\textsubscript{hmatrix} & parity matrix + wrapper opts  & $\approx$1{,}161  & matrix expander \\
L\textsubscript{cfg}     & 8-field configuration         & 69                & full generator \\
\bottomrule
\end{tabular}
\end{table}

\section{Method}\label{sec:method}
\textbf{Training regimes.} We evaluate seven. \emph{Zero-shot} and \emph{few-shot}
($k\in\{1,3,5\}$) use a base coder model with no weight updates. \emph{Retrieval-augmented
generation} supplies the $k$ nearest configurations as in-context demonstrations; we distinguish
it from a \emph{nearest-neighbour-copy} probe that returns the retrieved reference unchanged and,
under width-disjoint splits, is structurally incapable of succeeding (a memorization control).
\emph{LoRA} and \emph{full} supervised fine-tuning train on the labelled corpus;
\emph{continued-pre-training-then-SFT} first adapts on unlabelled RTL; \emph{verifier-guided RL}
optimizes a reward from compile, simulation, and formal-equivalence signals minus an area
penalty---a dense, perfect, cost-free reward most RTL settings lack.

\textbf{Models.} The fine-tuning rows use open-weight, self-hostable, Apache-2.0-licensed models:
Qwen2.5-Coder-32B-Instruct (dense) and Qwen3-Coder-30B-A3B-Instruct (a mixture-of-experts model
with $\approx$3B active parameters); the small full-fine-tune, CPT, and RL rows use
Qwen2.5-Coder-1.5B-Instruct. The from-scratch models are decoder-only transformers swept at
10M/50M/125M/350M parameters; the headline configuration is a sub-50M model that emits only the
eight-field configuration and runs on CPU, letting the deployed tool generate without a GPU.

\textbf{Energy methodology.} GPU energy is read from the device's firmware energy counter as a
delta across the run (avoiding power-polling error) and \emph{per GPU}, so a co-resident job does
not contaminate it---a subtlety that, uncorrected, roughly doubled our initial figures. We subtract
a multi-minute idle baseline and repeat each measurement ($n{=}3$). Clocks were unpinned and RAPL
unreadable, so figures are GPU energy only. We report joules per generated design and---the
deployment-relevant unit---joules per \emph{correct} design (J/design $\div$ pass rate).

\section{Results}\label{sec:results}
\textbf{Representation dominates.} Table~\ref{tab:t2} is the study's spine: pass rate for each
training regime $\times$ output representation. Read it two ways. \emph{Across the ladder} (the
few-shot row, the only regime run at every representation): every representation closer to
RTL---L\textsubscript{raw}, L\textsubscript{param}, L\textsubscript{hmatrix}---collapses to 0.17,
while L\textsubscript{cfg} reaches 0.83 with no training. \emph{Down the configuration column}:
every regime lands in a 0.89--1.00 band. Once the model emits the configuration, the training
method barely matters. The lever is the representation, not the optimizer.

\begin{table}[t]
\caption{Pass rate by regime $\times$ representation, scored by \textbf{L3 formal equivalence to
the reference} (in-distribution). ``---'': intentionally not run (Sec.~\ref{sec:whyblank}).}
\label{tab:t2}
\centering
\begin{tabular}{l cccc}
\toprule
Regime $\backslash$ Repr. & L\textsubscript{raw} & L\textsubscript{param} & L\textsubscript{hmatrix} & L\textsubscript{cfg} \\
\midrule
few-shot      & 0.17 & 0.17 & 0.17 & 0.83 \\
nn-copy       & ---  & ---  & ---  & 0.00 \\
RAG           & ---  & ---  & ---  & 1.00 \\
LoRA-SFT      & ---  & ---  & 0.00$^\dagger$ & 1.00 \\
full-SFT      & ---  & ---  & ---  & 0.975 \\
CPT-SFT       & ---  & ---  & ---  & 1.00 \\
verifier-RL   & ---  & ---  & ---  & 1.00 \\
from-scratch  & ---  & ---  & ---  & 0.887 \\
\bottomrule
\end{tabular}
\end{table}

\begin{figure}[t]\centering
\includegraphics[width=0.98\columnwidth]{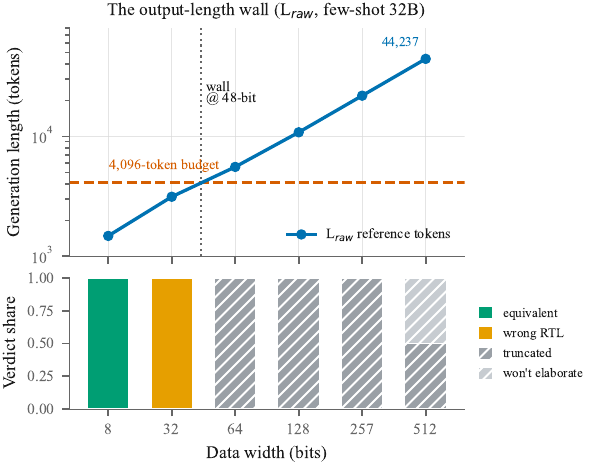}
\caption{Pass rate versus generation length, per width. Truncation is shown distinct from
incorrect RTL. Flat L\textsubscript{raw} generation is feasible only up to 32-bit words and hits the
wall at 48 bits (4{,}096-token budget).}\label{fig:wall}
\end{figure}

\textbf{The token wall.} The raw-RTL route does not merely lose accuracy; above a measurable width
it becomes infeasible. A 512-bit flat encoder is thousands of parity terms; the full design is
$\approx$44{,}237 tokens against 69 for the configuration, exceeding any current single-response
generation budget. Fig.~\ref{fig:wall} plots pass rate against generation length per width, with
truncation (``ran out of budget'') held as a verdict distinct from incorrect RTL (``got it
wrong''): the former is the wall, the latter a correctness failure, and conflating them would
erase the finding. At a 4{,}096-token budget, flat L\textsubscript{raw} generation is feasible only
up to \textbf{32-bit} data words ($\approx$3{,}137 tokens) and infeasible at 48 bits and beyond
($\approx$4{,}521 tokens). The wall is a property of the representation and the budget, not of the
model.

\textbf{Representation dominates even without a length wall.} The parity-matrix rung
(L\textsubscript{hmatrix}, $\le$1{,}161 tokens) is compact, so its failure is not a length wall but
a correctness wall. Five LoRA fine-tuning seeds do not close it: every seed reaches L3~$=0.00$, and,
checked reference-independently by exhaustive fault injection, \textbf{0 of 60} emitted matrices is a
valid SECDED code of \emph{any} dimension---the model over-generates the column list, emitting
$r\approx$26--181 check bits where 5--9 are required, often with even-weight or duplicate columns, so
single-bit errors are mis-corrected. It does not emit a valid-but-different code; it does not emit a
valid code at all. Only L\textsubscript{cfg}, where the oracle emits the matrix, reaches 1.00.
Training does not buy a valid parity matrix at this budget; the representation does.

\begin{figure}[t]\centering
\includegraphics[width=0.98\columnwidth]{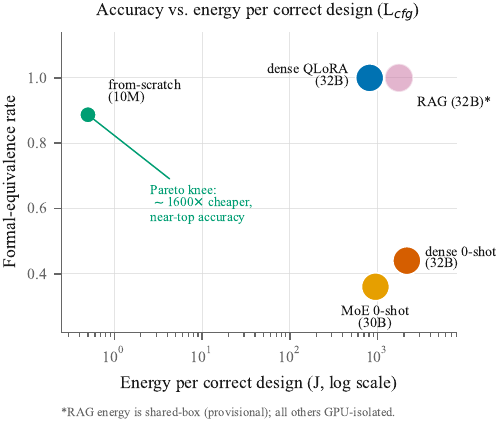}
\caption{Formal-equivalence rate versus energy per \emph{correct} design (log $x$-axis); bubble
size is parameter count, colour is representation. All points are GPU-isolated (NVML counter,
idle-subtracted, unpinned) except RAG, a provisional shared-box measurement (flagged in the plot).
The from-scratch configuration model occupies the high-accuracy, low-energy corner---the
specification-compression result made visual.}\label{fig:pareto}
\end{figure}

\textbf{Energy: the specification-compression result.} Fig.~\ref{fig:pareto} and
Table~\ref{tab:jcorrect} report formal-equivalence rate against joules per \emph{correct} design
(energy per design divided by pass rate): the from-scratch configuration model is roughly three
orders of magnitude ($\sim$1600$\times$) more efficient per correct design than the 32B fine-tune,
at comparable accuracy. This is the result in one number---emitting 69 tokens of
configuration into a verified generator, rather than tens of thousands of tokens of Verilog, is
where the efficiency comes from. Amortized training energy is small against lifetime inference:
0.032~kWh of fine-tuning over the deployment grid is $\sim$32~J/design, against $\sim$814~J/design
at inference for the same model---inference dominates amortized training by roughly 25$\times$, and
more as designs accumulate. This inverts the usual ML framing that fixates on training cost: lifetime cost is set by output length, which is the representation.

\begin{table}[t]
\caption{Joules per \emph{correct} design ($=$ J/design $\div$ pass rate). All rows are
\textbf{GPU-isolated} (per-GPU NVML energy counter, idle-subtracted, $n{=}3$, GPU energy only);
clocks unpinned (Sec.~\ref{sec:threats}).}
\label{tab:jcorrect}
\centering
\begin{tabular}{l r r r}
\toprule
Method (L\textsubscript{cfg}) & Pass & J/design & J/correct \\
\midrule
from-scratch ($<$50M) & 0.887 & 0.44 & 0.50 \\
MoE zero-shot         & 0.36  & 341  & 947 \\
dense QLoRA           & 1.00  & 814  & 814 \\
dense zero-shot       & 0.44  & 950  & 2{,}159 \\
\bottomrule
\end{tabular}
\end{table}

\textbf{Scaling, OOD, and refusal.} A from-scratch scaling sweep (10M--350M) with tokenizer and
grammar-constrained-decoding ablations confirms that a sub-50M configuration model suffices.
Out-of-distribution testing (Table~\ref{tab:ood}) separates the regimes that tie at 1.00
in-distribution. Unseen \emph{widths} degrade retrieval and the from-scratch model (0.89, 0.66)
while the pretrained fine-tunes hold; unseen option \emph{combinations} are easy for all
(0.93--1.00), so the width axis is where generalization breaks. The sharpest separator is
\emph{refusal of illegal configurations}: every pretrained-LLM regime refuses perfectly (1.00)---it
emits the illegal configuration faithfully and the oracle rejects it---but the efficiency-winning
grammar-constrained from-scratch model refuses only \textbf{0.59}, silently legalizing 41\% of
illegal specs. Trained only on the legal manifold, its constrained decoder cannot emit an invalid
config, so it rounds illegal requests to nearby valid ones. The cheapest model carries a safety
cost the expensive ones do not---an efficiency--safety trade-off only an illegal-input test exposes.

\begin{table}[t]
\caption{Out-of-distribution behaviour (L\textsubscript{cfg}); config-exact, except the last column
(refusal accuracy on illegal specifications: fraction \emph{not} silently legalized).}
\label{tab:ood}
\centering
\begin{tabular}{l cccc}
\toprule
Regime & \makecell{ID\\(deploy)} & \makecell{OOD\\width} & \makecell{OOD\\combo} & \makecell{illegal\\refusal} \\
\midrule
RAG          & 1.00  & 0.887 & 0.93 & 1.00 \\
LoRA-SFT     & 1.00  & 1.00  & 1.00 & 1.00 \\
CPT-SFT      & 1.00  & 1.00  & 1.00 & 1.00 \\
verifier-RL  & 1.00  & 0.975 & 0.95 & 1.00 \\
from-scratch & 0.887 & 0.662 & 1.00 & \textbf{0.59} \\
\bottomrule
\end{tabular}
\end{table}

\textbf{Why the raw-RTL cells are blank.}\label{sec:whyblank} The empty cells in Table~\ref{tab:t2}
are not omissions: one does not fine-tune a model to emit 44k tokens of flat Verilog, so the
L\textsubscript{raw}/L\textsubscript{param} cells are blank because the representation is
infeasible. The one compact rung without a length wall, L\textsubscript{hmatrix},
we deliberately filled: the gap there is correctness, not feasibility.

\section{Discussion}\label{sec:discussion}
\textbf{Don't fine-tune; retrieve.} Four regimes reach 1.00 on the configuration, and one of
them---retrieval-augmented generation---requires no weight updates, negligible energy, and trivial
updating when the design space changes. On a bounded, template-structured task the honest
recommendation is therefore not ``fine-tune a model'' but ``retrieve and let the verified back-end
expand.'' The OOD table stress-tests this and it largely survives: retrieval refuses illegal
configurations perfectly (1.00), and its one weakness is unseen widths (0.89), where a light
fine-tune buys extrapolation. Notably the regime that \emph{does} fail the safety test is the
ultra-cheap grammar-constrained from-scratch model---so ``cheapest'' and ``safest'' are not the same
choice.

\textbf{When is an LLM the right EDA tool?} Not to replace a working deterministic generator on its
own turf---the incumbent shell command is faster and always correct---but at the edges it cannot
serve: mapping an underspecified or natural-language request onto a valid configuration, behind a
verification gate that returns RTL only when it is proven equivalent to the reference. The deployed
architecture reflects this: the model writes the interface, the generator writes the implementation,
and the oracle decides.

\section{Threats to Validity}\label{sec:threats}
We study one ECC scheme and one generator, so external validity is limited; the mechanism---an
output-length wall and a representation that dodges it---is not specific to ECC, but a second
oracle-complete task family is left to future work. The reference generator emits
template-structured RTL, so results may not transfer to unconstrained human-written RTL; we quantify
corpus entropy (few distinct encoders/decoders) so readers can calibrate. ``Correctness'' is
equivalence to a standard minimal-Hsiao reference; a different valid construction would define a
different reference, but a freshly-generated wrapper is internally self-consistent regardless. The
energy figures are GPU-isolated---which removed a real co-resident-job contamination that had
inflated earlier numbers---but the clocks are \emph{unpinned} and RAPL was unreadable, so they are
GPU energy only. The \emph{relative} orderings (configuration $\ll$ raw RTL; MoE $<$ dense; inference
$\gg$ training) are robust to all of this and are the claims we rest on. The three OOD columns come
from a single run per regime (reproducing the in-distribution headline exactly for three regimes and
within seed variance for the other two), so the four-way tie is three robust regimes plus one
higher-variance. Natural-language specifications are model-paraphrased; a held-out set of
\emph{real} user prompts, and a systematic contamination probe against pre-training data, are left
to future work.

\section{Conclusion}\label{sec:conclusion}
On a bounded, formally-specifiable hardware-generation task with a decidable correctness oracle, the
output representation dominates the training regime: raw RTL generation hits a hard output-length
wall and becomes infeasible above a measurable width, while emitting a short configuration into a
verified generator reaches near-perfect accuracy under essentially any training regime---including
none. A sub-50M from-scratch model, backed by the generator, reaches 0.89 formal-equivalence against
a 32B fine-tune's 1.00 at $\sim$1600$\times$ lower energy per correct design, and inference energy
dominates amortized training. The engineering lesson is to learn the interface, not the
implementation; the methodological lesson is that template-structured RTL benchmarks must control
for cross-split near-duplication. We release the generator, the oracle-complete benchmark, the
verification harness, and the from-scratch weights.

\section*{Reproducibility}
All code, the frozen and hashed data splits, the verification harness, and the from-scratch weights
are released in a public repository at \url{https://github.com/thesid16/learn-the-interface}. Every
number in this paper is recorded---with its verification level, sample size, and provenance---in a
machine-readable results file at the repository root, and the paper cites no number absent from it.
A clean checkout reproduces the generator and oracle (property-tested), byte-level determinism
(every configuration identical across two runs), and the width-disjoint leakage audit.

\section*{Acknowledgment (AI Disclosure)}
In accordance with the conference policy: an AI coding assistant was used for portions of the software implementation. All numerical results were produced by the released, deterministic pipeline and verified by formal equivalence; the authors are responsible for all claims.

\bibliographystyle{IEEEtran}

\end{document}